\documentclass[journal]{IEEEtran}
\ifCLASSINFOpdf
\else
\fi
\usepackage[pdftex]{graphicx}
\usepackage{multirow}

\begin{document}
%
\title{Image Companding and Inverse Halftoning using Deep Convolutional Neural Networks}
%
%
%

%
\author{Xianxu~Hou and Guoping~Qiu

\thanks{X. Hou is with the School of Computer Science, University of Nottingham Ningbo China and a visiting student in the College of Information Engineering, Shenzhen University, China. email: xianxu.hou@nottingham.edu.cn}
\thanks{G. Qiu is with the College of Information Engineering, Shenzhen University, China and with the School of Computer Science, the University of Nottingham, United Kingdom. email: qiu@szu.edu.cn and guoping.qiu@nottingham.ac.uk}}
\maketitle

\begin{abstract}
This paper presents a deep learning technology for tackling two traditional low-level image processing problems, companding and inverse halftoning. This paper makes two main contributions. First, to the best knowledge of the authors, this is the first work that has successfully developed deep learning based solutions to these two traditional low-level image processing problems. As well as introducing new methods to solving well-known image processing problems, this paper also contributes to a growing literature that demonstrates the power of deep learning in solving traditional signal processing problems. Second, building on insights into the properties of visual quality of images and the internal representation properties of a deep convolutional neural network (CNN) and inspired by recent success of deep learning in other image processing applications, this paper has developed an effective deep learning method that trains a deep CNN as a nonlinear transformation function to map a lower bit depth image to higher bit depth or from a halftone image to a continuous tone image, and at the same time employs another pretrained deep CNN as a feature extractor to derive visually important features to construct the objective function for the training of the transformation CNN. Extensive experimental results are presented to show that the new deep learning based solution significantly outperforms previous methods and achieves new state-of-the-art results.

\end{abstract}

\begin{IEEEkeywords}
Image Companding, Inverse Halftoning, CNNs, Perceptual Loss.
\end{IEEEkeywords}

%
\IEEEpeerreviewmaketitle

\begin{figure*}
\centering
  \includegraphics[width=\textwidth]{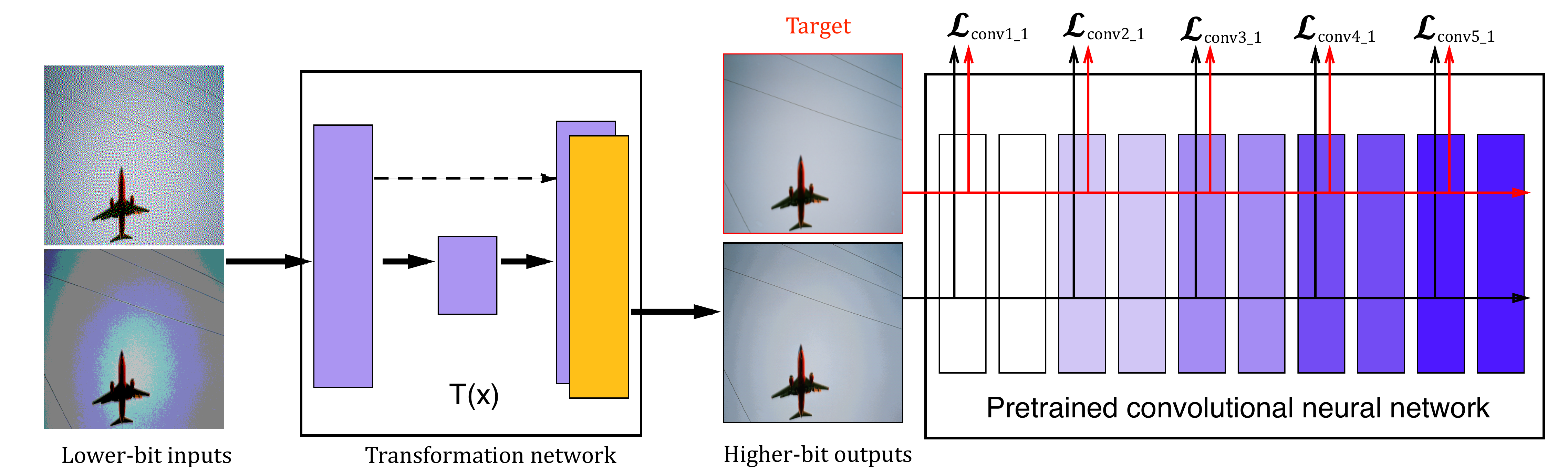}
  \caption{Method overview. A transformation convolutional neural network (CNN) to expand lower-bit images to higher-bit ones. A pretrained deep CNN for constructing perceptual loss to train the transformation network.}
  \label{fig:overview}
\end{figure*}

\section{Introduction}
\IEEEPARstart{C}{ompanding} is a process of compression and then expanding, allowing signals with a higher dynamic range to be transmitted with a lower dynamic range by reducing the number of bits. This technique is widely used in telecommunication and signal processing such as audio processing. For image processing, companding could be regarded as an encoding and decoding framework. The encoding or quantized compression process, while fairly simple and efficient, could also produce a lot of undesirable artifacts, such as blocking artifacts, contouring and ringing effects (see Fig. \ref{fig:overview}). These degraded artifacts become more obvious with lower bit quantization.

Inverse halftoning, another similar image processing problem considered in this paper, is the inversed process for halftoning. Halftone images are binary images served as analog representation and widely used in digital image printing, trying to convey the illusion of having a higher number of bit levels (continuous-tone) to maintain the overall structure of the original images. As a result, distortions will be introduced to the halftone images due to a considerable amount of information being discarded. Inverse halftoning, on the other hand, addresses the problem of recovering a continuous-tone image from the corresponding halftoned version. This inversed process is needed since typical image processing techniques such as compression and scaling can be successfully applied to continuous-tone images but very difficult to halftone images.

However, these two problems are ill-posed considering that there could be an infinite number of possible solutions. They are essentially one-to-many mappings and the input image could be transformed into an arbitrary number of plausible outputs even if the compression and halftone methods are known in advance. Solving both problems requires to find a way to estimate and add more information into the images that do not exist. There are no well-defined mathematic functions or guidelines to describe the mappings to produce high-quality images. 

In this paper, we take advantage of the recent development in machine learning, in particular deep convolutional neural networks (CNNs), which have become the state-of-the-art workforce for most computer vision tasks \cite{krizhevsky2012imagenet,simonyan2014very}. Unlike previous human-engineered methods \cite{li2005compressing,kite2000fast,shen2001robust,easley2009inverse}, we formulate the two image processing problems, i.e., companding and inverse halftoning, from the perspective of machine learning. We train deep convolutional neural networks as non-linear mapping functions in a supervised manner to expand images from a lower bit depth to a higher bit depth to reduce artifacts in image companding and to produce continuous-tone images in inverse halftoning. Moreover, we also investigate the effect to construct loss functions based on different level convolutional layers, which have shown different properties when applying an inverting processing to reconstruct the encoded images \cite{mahendran2015understanding}.

Our core contributions in this work are two folds. Firstly, to the best knowledge of the authors, this is the first work that has successfully developed deep learning based solutions to these two traditional image processing problems. This not only introduces new methods to tackle well-known image processing problems but also contributes to the literature that demonstrates the power of deep learning in solving traditional signal processing problems. Secondly, building on insights into the properties of visual quality of images and the hidden representation properties of deep CNNs, and also inspired by recent works that use deep CNNs in other image processing applications \cite{gatys2015neural,johnson2016perceptual,hou2017deep}, we take full advantage of the convolutional neural networks both in the nonlinear mapping functions and in the neural networks loss functions for low-level image processing problems. We not only use a deep CNN as a nonlinear transformation function to map a low bit depth image to a higher bit depth image or from a halftone image to a continuous tone image, but also employ another pre-trained deep CNN as a feature extractor or convolutional spatial filter to derive visually important features to construct the objective function for the training of the transformation neural network. Through these two low-level image processing case studies, we demonstrate that a properly trained deep CNN can capture the spatial correlations of pixels in a local region and other visually important information, which can be used to help a deep CNN to infer  the ``correct" values of pixels and their neighbors. Our work further demonstrates that halftone images and heavily compressed low bit depth images, even though showing visually annoying artifacts, they have preserved the overall structures of the images which are sufficient to enable deep neural networks to recover the original signals to a high degree of fidelity. 

\section{Related Works}

\subsection{Image Companding}
Companding, a combination of the words \textbf{com}pressing and ex\textbf{panding}, is a signal processing technique to allow signals with a large dynamic range transmitted in a smaller dynamic range format. This technique is widely used in digital telephony systems. Image companding \cite{yang2004integer,bhooshan20102d,li2005compressing} is designed to squeeze higher-bit images to lower bit ones, based on which to reproduce outputs with higher bits. Multi-scale subband architecture \cite{li2005compressing} successfully compressed high dynamic range (HDR) images to displayable low dynamic range (LDR) ones. They also demonstrated that the compression process can be inverted by following the similar scheme as the previous compression. As a result, low dynamic range images can be expanded to approximate the original higher-bit ones with minimal degradation.

\subsection{Halftoning and Inverse Halftoning}
The typical digital halftoning process is considered as a technique of converting a continuous-tone grayscale image with 255 color levels (8 bits) into a binary black-and-white image with only 0 and 1 two color levels (1 bit). These binary images could be reproduced to ``continuous-tone" images for humans based on an optical illusion that tiny dots are blended into smooth tones by human eyes at a macroscopic level. In this work, we focus on the most popular halftoning technique known as error diffusion, in which the residual quantization error of a pixel is distributed to neighboring pixels. Floyd–Steinberg dithering is commonly used by image manipulation software to achieve error diffused halftoning based on a simple kernel. The reversed processing known as inverse halftoning is to reconstruct the continuous-tone images from halftones. Many approaches to addressing this problem have been proposed in the literature, including non-linear filtering \cite{shen2001robust}, vector quantization \cite{ting1994error}, projection onto convex sets \cite{unal2001restoration}, MAP projection \cite{stevenson1997inverse}, wavelets-based \cite{neelamani2002winhd}, anisotropic diffusion \cite{kite2000fast}, Bayesian-based \cite{liu2011inverse}, a method by combining low-pass filtering and super-resolution \cite{minami2012inverse}, Look-up table \cite{mese2001look}, sparse representation \cite{son2012inverse}, local learned dictionaries \cite{son2014local} and coupled dictionary training \cite{freitas2016enhancing}.

\subsection{Deep Learning for Image Transformation}
In this work, we seek to formulate the image companding and inverse halftoning as image transformation problems and employ deep convolutional neural networks as non-linear functions to map input images to output images for different purposes. Recent deep CNNs have become a common workhorse behind a wide variety of image transformation problems. These problems can be formulated as per-pixel classification or regression by defining low level loss. Semantic segmentation methods \cite{long2015fully,eigen2015predicting,noh2015learning} use fully convolutional neural networks trained by per-pixel classification loss to predict dense scene labels. End-to-end automatic image colorization techniques \cite{iizuka2016let,larsson2016learning} try to colorize grayscale image based on low level losses. Other works for depth \cite{eigen2014depth,liu2015deep} and edge detection \cite{xie2015holistically} are also similar to transform input images to meaningful output images through deep convolutional neural networks, which are trained with per-pixel classification or regression loss. However the per-pixel measurement essentially treats the output images as ``unstructured" in a sense that each pixel is independent with all other pixels for a given image.

\begin{figure*}
\centering
  \includegraphics[width=\textwidth]{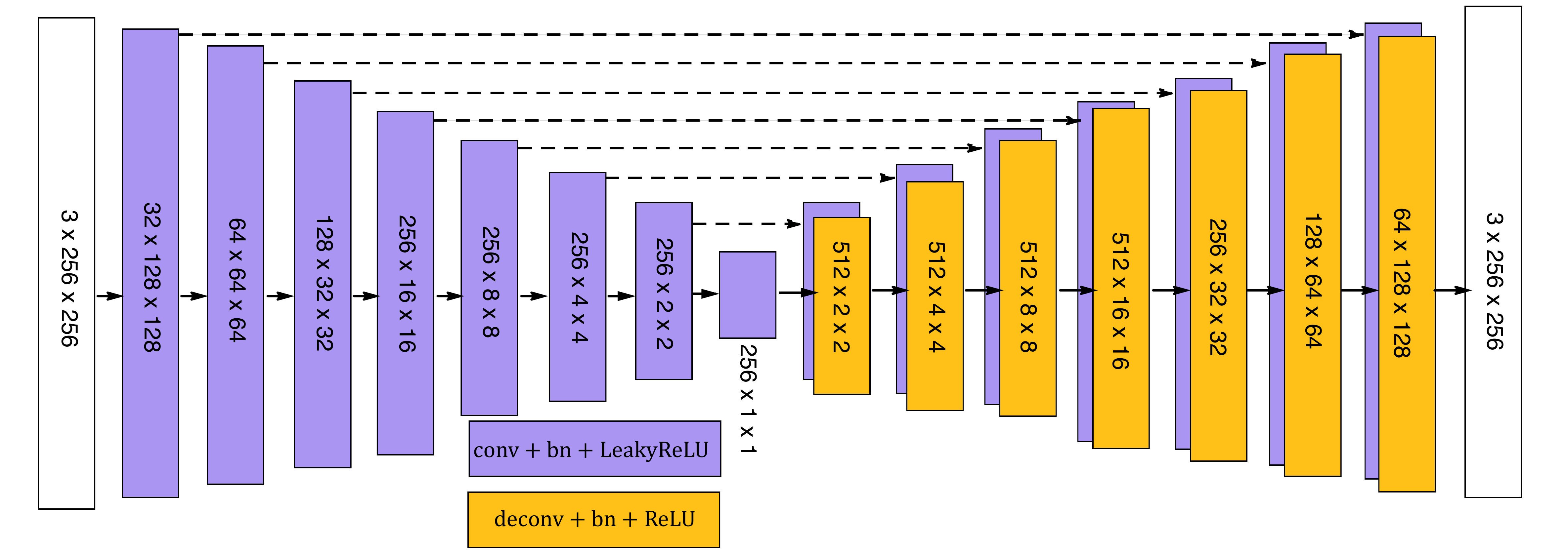}
  \caption{The architecture of the transformation networks. The ``U-Net" network is an encoder-decoder with skip connections between the encoder and decoder. The dash-line arrows indicate the features from the encoding layers are directly copied to the decoding layers and form half of the corresponding layers' features.}
  \label{fig:architecture}
\end{figure*}

Considering the shortcoming of per-pixel loss, other ``structured" measurements have been proposed such as structural similarity index measure (SSIM) \cite{wang2004image} and conditional random fields (CRF) \cite{chen2014semantic}, which take context into account. These kinds of ``structured" loss have been successfully applied to different image transformation problems. However these measurements are human-crafted, the community has successfully developed structure loss directly learned from images. Generative adversarial networks (GANs) \cite{goodfellow2014generative} are able to generate high-quality images based on adversarial training. Many works have tried to apply GANs in conditional settings such as discrete labels \cite{mirza2014conditional}, texts \cite{reed2016generative} and of course images. Image-conditioned GANs involve style transfer \cite{li2016precomputed}, inpainting \cite{pathak2016context}, frame prediction \cite{mathieu2015deep}. In addition, image-to-image translation framework \cite{isola2016image} based on adversarial loss can effectively synthesize photos under different circumstances.

Another way to improve per-pixel loss is to generate images by optimizing a perceptual loss which is based on high level features extracted from pretrained deep convolutional neural networks. By optimizing individual deep features \cite{yosinski2015understanding} and maximizing classification score \cite{simonyan2013deep}, images can be generated for a better understanding of hidden representations of trained CNNs. By inverting convolutional features \cite{dosovitskiy2016inverting}, the colors and the rough contours of an image can be reconstructed from activations in pretrained CNNs. In addition, artistic style transfer \cite{gatys2015neural} can be achieved by jointly optimizing the content and style reconstruction loss based on deep features extracted from pretrained CNNs. A similar method is also used for texture synthesis \cite{gatys2015texture}. Similar strategies are also explored to achieve real-time style transfer and super-resolution \cite{johnson2016perceptual}. Deep feature consistent variational autoencoder \cite{hou2017deep} is proposed to generate sharp face images and manipulate facial attributes by minimizing the difference between the deep features of the output images and target images.

\section{Method}
In this work, we propose to use deep convolutional neural networks with skip connections as non-linear mapping functions to expand images from a lower bit depth to a higher bit depth. The objective of generating the higher bit depth version of the image is to ensure that this image is visually pleasing and to capture the essential and visual important properties of the original version of the image. Instead of using per-pixel losses, i.e. measuring pixel-wise difference between the output image and its target (the original) image, we measure the difference between the output image and target image based on the high level features extracted from pretrained deep convolutional neural networks. The key insight is that the pretrained networks have already encoded perceptually useful information we desired, such as the spatial relationship between pixels nearby. Our system is diagrammatically illustrated in Fig. \ref{fig:overview}, which consists of two parts: an autoencoder transformation neural network $T(x)$ to achieve end-to-end mapping from an input image to an output image, and a pretrained neural network $\Phi(x)$ to define the loss function.

\subsection{Network Architecture}
Our non-linear mappings are deep convolutional neural networks, which have been demonstrated to have state-of-the-art performances in many computer vision tasks. Successful network architecture like AlexNet \cite{krizhevsky2012imagenet}, VGGNet \cite{simonyan2014very} and ResNet \cite{he2016deep} are designed for high level tasks like image classification to output a single label, and they cannot be directly applied to image processing problems. Instead, previous works have employed an encoder-decoder architecture \cite{hou2017deep,johnson2016perceptual} to firstly encode the input images through several convolutional layers until a bottleneck layer, followed by a reversed decoding process to produce the output images. Such encoder-decoder architecture forces all the information to pass through the networks layer by layer. Thus the final generated images are produced by higher layers' features. However for image processing, the output images can retain a great deal of lower layers' information of the input images, and it would be better to incorporate lower layers features in the decoding process. Based on the architecture guidelines of previous work on image segmentation \cite{ronneberger2015u}, image-to-image translation \cite{isola2016image} and DCGAN \cite{radford2015unsupervised}, we add skip connections to construct a ``U-Net" network to fuse lower layers and higher layers features and employ fully convolutions for image transformation.

The details of our model are shown in Fig. \ref{fig:architecture}, we first encode the input image to lower dimension vector by a series of stride convolutions, which consists of 4 x 4 convolution kernels and 2 x 2 stride in order to achieve its own downsampling. We also use a similar approach for decoding to allow the network to learn its own upsampling by using deconvolutions \cite{long2015fully}. Spatial batch normalization \cite{ioffe2015batch} is added to stabilize the deep network training after each convolutional layer except the input layer of the encoder and the last output layer of decoder as suggested in \cite{radford2015unsupervised}. Additionally leaky rectified activation (LeakyReLU) and ReLU are served as non-linear activation functions for encoder and decoder respectively. Finally we directly concatenate all the encoding activations to the corresponding decoding layers to construct a symmetric ``U-Net" structure \cite{ronneberger2015u} to fuse the features from both low layers and high layers.

\begin{figure*}
\centering
  \includegraphics[width=\textwidth]{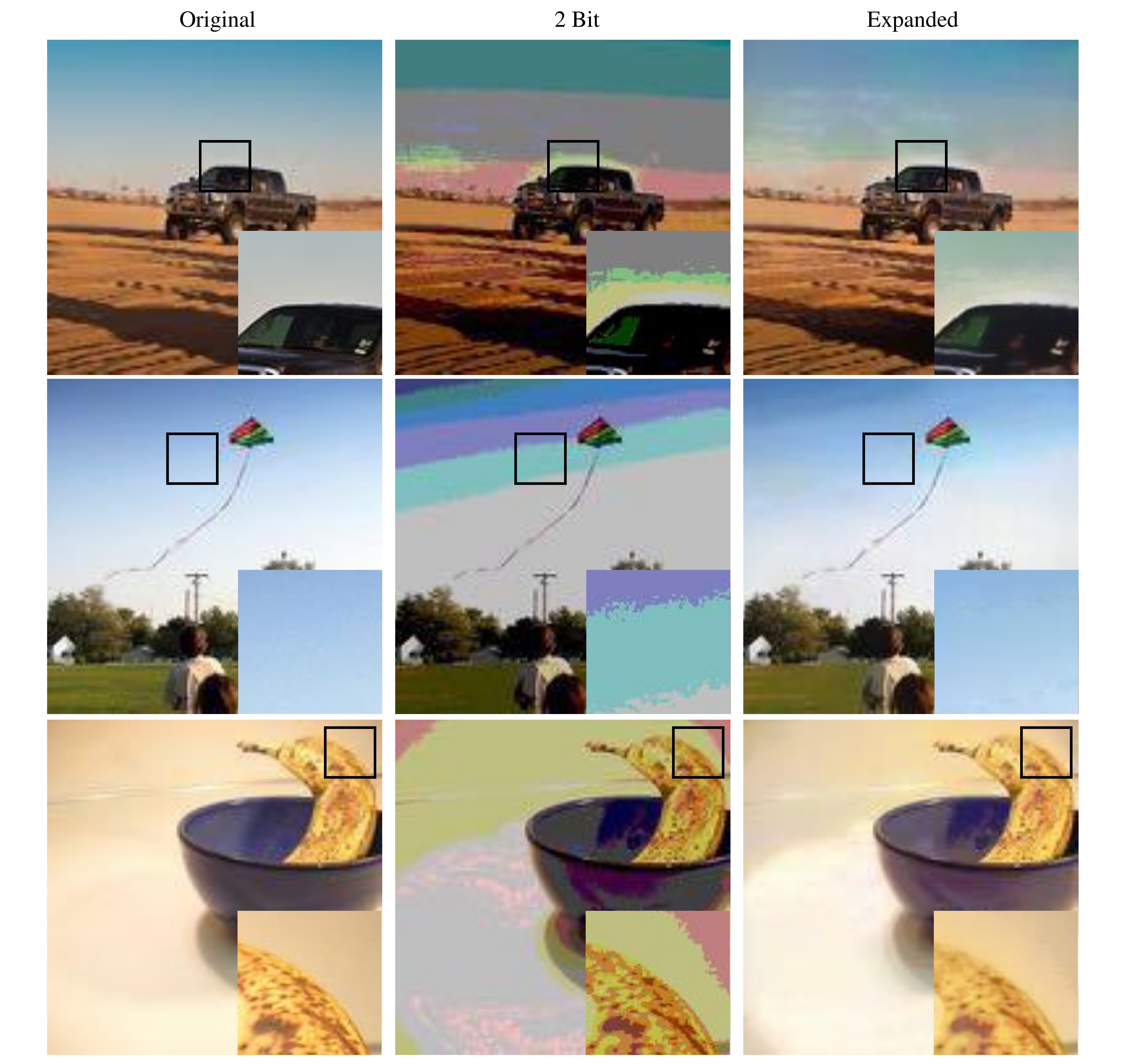}
  \caption{Results on color images from Microsoft COCO validation split for blocking and contour artifacts reduction. A pair of compressed 2 bit images and the corresponding expanded ones are shown together. Additionally an enlarged sub-image of each image is given at the bottom for better comparison.}
  \label{fig:depths_companding1}
\end{figure*}

\begin{figure*}
\centering
  \includegraphics[width=\textwidth]{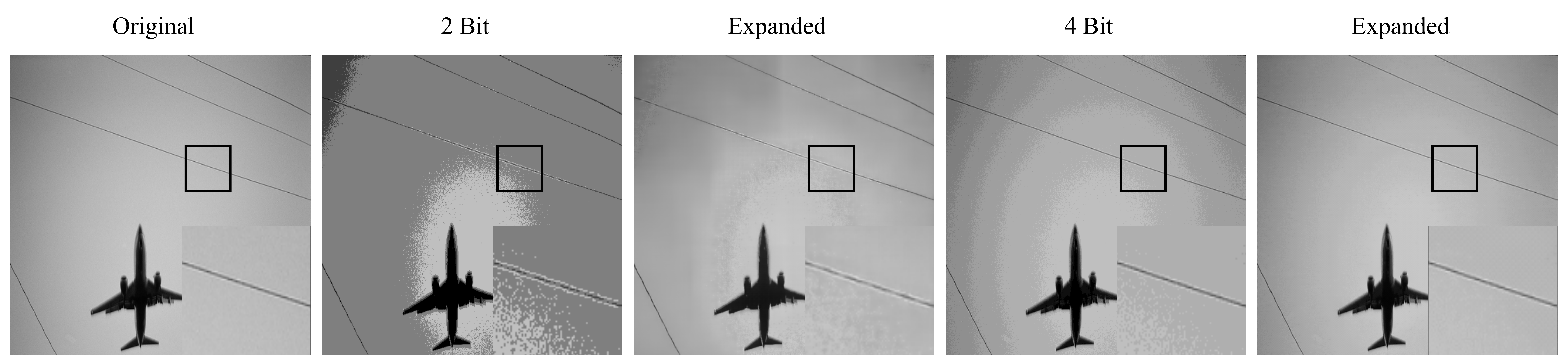}
  \caption{Results on grayscale images from Microsoft COCO validation split for blocking and contour artifacts reduction. A pair of compressed 2 bit and 4 bit images and the corresponding expanded ones are shown together. Additionally an enlarged sub-image of each image is given at the bottom for better comparison.}
  \label{fig:depths_companding2}
\end{figure*}

\begin{figure}[!tb]
  \centering
  \includegraphics[width=9cm]{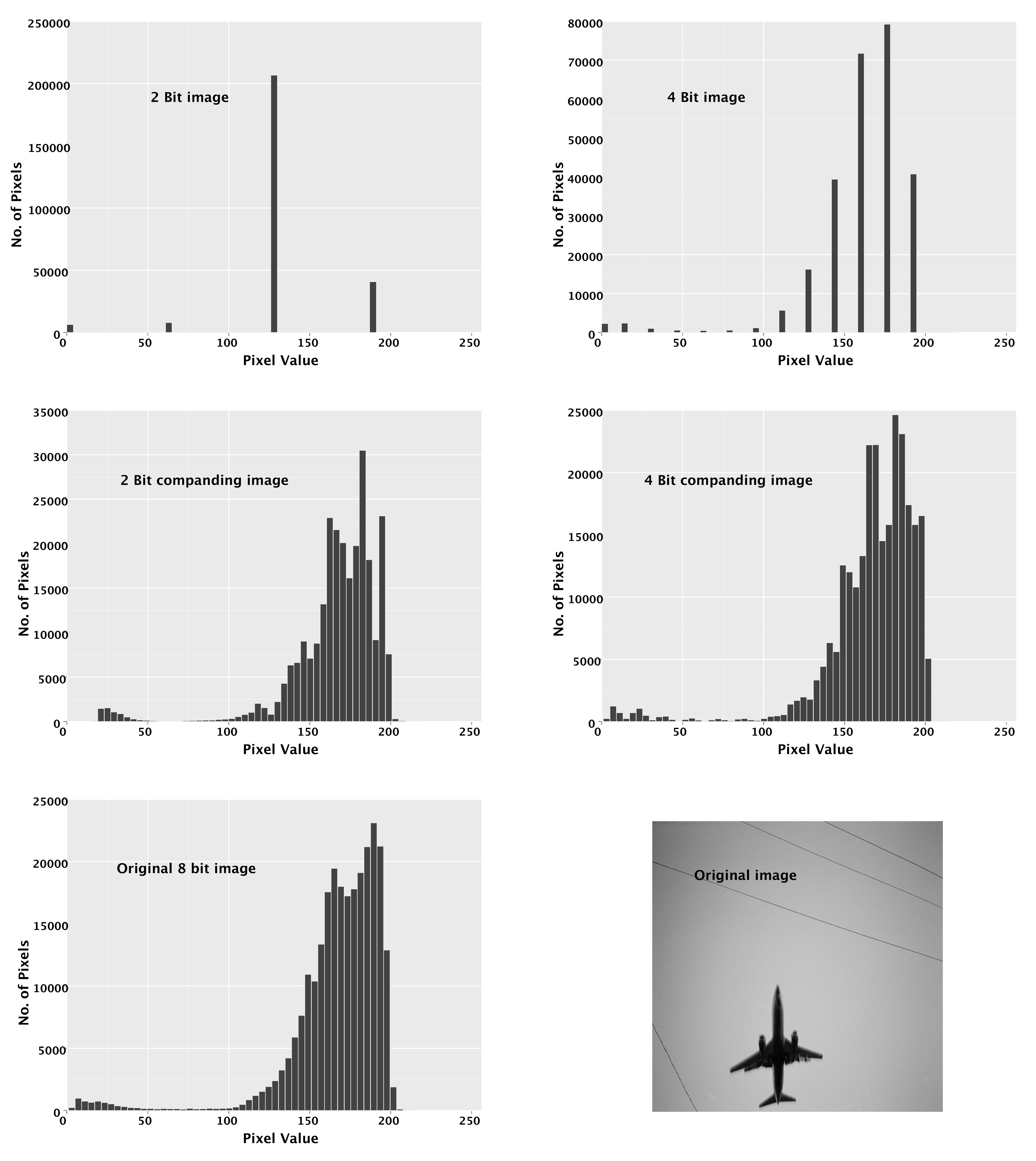}
  \caption{Intensity histogram of different compressed and expanded images.}
  \label{fig:hist_companding}
\end{figure}

\begin{figure*}[!tb]
  \centering
  \includegraphics[width=17cm]{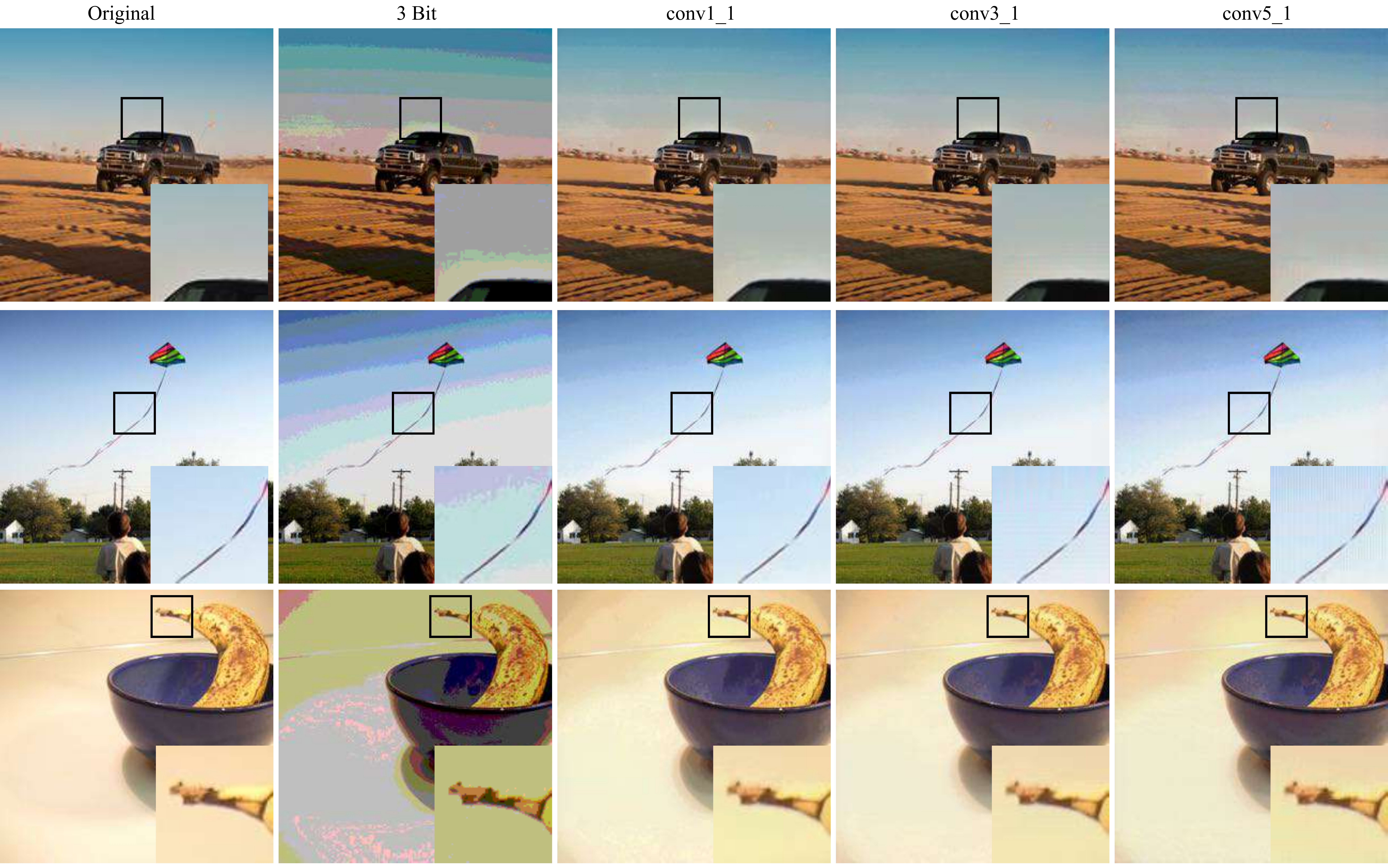}
  \caption{Results on color images for blocking and contour artifacts reduction. The compressed images are fixed to 3 bits with 8 color levels for each channel. The conv1\_1, conv3\_1, conv5\_1 are the expanded results produced by the models trained with perceptual loss constructed by corresponding convolutional layers. Additionally an enlarged sub-image of each image is given at the bottom for better comparison.}
  \label{fig:depth3_companding}
\end{figure*}

\subsection{Perceptual Loss}
It is well known that per-pixel loss for regression and classification is problematic and could produce blurry outputs or other visual artifacts. This is because each pixel is regarded as an individual object for optimization, resulting in average outputs to some degree. A better strategy is to construct the loss by incorporating the spatial correlation information. Rather than encouraging matching each individual pixels of input and output images, we follow previous works \cite{hou2017deep,johnson2016perceptual,gatys2015neural} to measure the difference between two images at various deep feature levels based on pretrained deep convolutional neural networks. We seek to capture the input images' spatial correlations by means of convolution operations in the deep CNNs.

We denote the loss function as $\mathcal{L}(\hat{y}, y)$ to measure the perceptual difference between two images. As illustrated in Fig. \ref{fig:overview}, both the output image $\hat{y} = T(x)$ generated by the transformation network and the corresponding target image $y$ are fed into a pretrained deep CNN $\Phi$ for feature extraction. We use $\Phi_i(y)$ to represent the hidden representations of image $y$ at $i^{th}$ convolutional layer. $\Phi_i(x)$ is a 3D array of shape [$C_i$, $W_i$, $H_i$], where $C_i$ is the number of filters, $W_i$ and $H_i$ are the width and height of the given feature map of the $i^{th}$ convolutional layer. The final perceptual loss of two images at $i^{th}$ layer is the Euclidean distance of the corresponding 3D arrays as following:

\begin{equation}
\footnotesize
\mathcal{L}_i(\hat{y}, y) =  \frac{1}{C_iW_iH_i}  \sum_{c=1}^{C_i}  \sum_{w=1}^{W_i}  \sum_{h=1}^{H_i} (\Phi_i(\hat{y})_{c,w,h} - \Phi_i(y)_{c,w,h})^2
\end{equation}

In fact, above loss still follows the per-pixel manner if we treat the hidden features which are 3D arrays as ``images" with more than 3 color channels. However this kind of loss has already incorporated the spatial correlation information because the ``pixels" in these images are the combinations of the original pixels through convolution operations.

\subsection{Training Details}
Our implementation uses open source machine learning framework Torch \cite{collobert2011torch7} and a Nvidia Tesla K40 GPU to speed up training. The pretrained 19-layer VGGNet \cite{simonyan2014very} is chosen as the loss network for deep feature extraction which is fixed during the training. In addition, due to the similar convolutional architecture, the loss network can be seamlessly stacked to our ``U-Net" neural network to achieve end-to-end training. The training images are of the shape 256$\times$256 and we train our model with a batch size of 16 for 30,000 iterations. Adam optimizer \cite{kingma2014adam} is used for stochastic optimization with a learning rate of 0.0002. For the LeakyReLU in the encoder, the slope of the leak is set to 0.2 in all layers. Additionally we experiment with conv1\_1, conv2\_1, conv3\_1, conv4\_1 and conv5\_1 layers in VGGNet to construct perceptual loss for comparison.

\section{Experimental Results}
In our experiments, we use Microsoft COCO dataset \cite{lin2014microsoft} which is a large-scale database containing more than 300,000 images as our training images. We resize the training images to 256$\times$256 as our inputs to train our models. We perform experiments on two image processing problems: image companding and inverse halftoning.

\subsection{Image Companding}
One essential part of image companding is to expand lower bit images to higher bit outputs. This technique has been investigated in the context of high dynamic range (HDR) imaging \cite{li2005compressing}, firstly compressing the range of an HDR image into an LDR image, at which point the process is then reversed to retrieve the original HDR image.

Since it is impossible to display a true HDR image with more than 8 bits, we use 8 bit images as our highest bit depth images in the experiments. The 8 bit images are reduced by different depths as the lower bit depth images, and then expanded back to 8 bits. Take 4 bit images for example, they can only have 16 different levels for each color channel while there are 256 different levels for 8 bit images. The default approach \cite{li2005compressing} for converting 8 bit images to 4 bit images is to divide by 16 to quantize the color level from 256 to 16, which will be then scaled up to fill the full range of the display. Mathematically we can use the formula below to easily convert 8 bit images to different lower bit outputs. This operation can be applied to both grayscale images and color images by processing each channel separately.

\begin{equation}
\footnotesize
I_{low} = \lfloor \frac{I_{high}}{2^{(h - l)}} \rfloor 2^{(h - l)}
\end{equation}

where the $I_{low}$ and $I_{high}$ are the pixel intensity of converted lower and higher bit depth images respectively, $l$ and $h$ are the bit depth for lower and higher bit depth images.

We first preprocess the training images to different lower-bit ones as input data, and use the original images as higher-bit targets we want to retrieve. After training, the validation split of Microsoft COCO is used for testing. We first compare the results of different lower-bit input images, and then evaluate how the perceptual loss constructed from different convolutional layers affects the expanding quality.

\begin{table}[]
\scriptsize
\centering
\caption{The average companding results of PSNR(dB) and SSIM for 100 color and grayscale images randomly selected from Microsoft COCO validation split. The expanded results were based on a perceptual loss constructed using conv1\_1 layer.}
\label{table:psnr_ssim_conv12_companding}
\begin{tabular}{c|c|c|c|c|c}
\hline
\multicolumn{2}{c|}{\multirow{2}{*}{\begin{tabular}[c]{@{}c@{}}Input Bit-depth\end{tabular}}} & \multicolumn{2}{c|}{PSNR} & \multicolumn{2}{c}{SSIM} \\ \cline{3-6}
\multicolumn{2}{c|}{}                                        & Compressed     & Expanded     & Compressed     & Expanded     \\ \hline
\multirow{2}{*}{Bit 1}               & Color                  & 11.73          & 18.67    & 0.40           & 0.55     \\ \cline{2-6}
                                     & Grayscale              & 11.58          & 18.81    & 0.35           & 0.49     \\ \hline
\multirow{2}{*}{Bit 2}               & Color                  & 17.37          & 25.65    & 0.67           & 0.81     \\ \cline{2-6}
                                     & Grayscale              & 17.29          & 25.84    & 0.61           & 0.74     \\ \hline
\multirow{2}{*}{Bit 3}               & Color                  & 23.13          & 30.79    & 0.85           & 0.90     \\ \cline{2-6}
                                     & Grayscale              & 23.16          & 31.33    & 0.78           & 0.87     \\ \hline
\multirow{2}{*}{Bit 4}               & Color                  & 29.03          & 34.52    & 0.94           & 0.95     \\ \cline{2-6}
                                     & Grayscale              & 29.19          & 36.69    & 0.90           & 0.94     \\ \hline
\multirow{2}{*}{Bit 5}               & Color                  & 34.85          & 37.59    & 0.98           & 0.97     \\ \cline{2-6}
                                     & Grayscale              & 35.08          & 40.24    & 0.96           & 0.97     \\ \hline
\end{tabular}
\end{table}

\subsubsection{Different Bit Depths}

We have separately trained models for different lower-bit input images for comparison and use the conv1\_1 layer of VGGNet to construct the perceptual loss for all the models.

\textbf{Qualitative Results.}
Fig. \ref{fig:depths_companding1} and Fig. \ref{fig:depths_companding2} show the qualitative results for a variety of color and grayscale images taken from Microsoft COCO 2014 validation split. We can see that the linearly quantized lower-bit images display severe blocking and contouring artifacts. The compression process amplifies low amplitudes and high frequencies which dominate the quantization artifacts because we try to show a lower dynamic range image on a higher dynamic range displayable device. For instance, our device is appropriate for the original 8 bit targets with 256 color levels. We could drop the bit depths of the original images by 5 bits and linearly quantize them to 3 bit images with only 8 color levels. Since the compressed images contain 5 fewer bits, they should be theoretically displayed on 1/32 dynamic range device. It is obvious that this kind of lossy compression introduces visible artifacts in pixel blocks and at block boundaries.

We also show the corresponding expanded images (Fig. \ref{fig:depths_companding1} and Fig. \ref{fig:depths_companding2}) retrieved from our models. The blocking and contouring artifacts are effectively reduced to show smooth appearance in the expanded outputs. For example in the airplane image in Fig. \ref{fig:depths_companding2}, the compressed images show obvious contouring artifacts in the sky while the expanded images have homogeneous gradually changing colors. And this can be further validated from the distribution of intensity histograms. Fig. \ref{fig:hist_companding} shows the intensity histograms for the compressed 2 and 4 bit airplanes and the expanded ones in Fig. \ref{fig:depths_companding2}. It is clear that our methods are able to infer the ``correct" values for a single pixel based on its neighbors, and convey a more attractive impression with rich and saturated colors.



\textbf{Quantitative Results.}
In order to have a comprehensive quantitative evaluation for our models, we report peak signal-to-noise ratio (PSNR) and structural similarity index measure (SSIM) \cite{wang2004image} for quality assessment. PSNR is per-pixel based measurement defined via the mean squared error (MSE) while SSIM index is known as perceptual-aware method for measuring the similarity between two images. For both measurements, a higher value indicates better quality. Table \ref{table:psnr_ssim_conv12_companding} summarizes the average PSNR (dB) and SSIM values of 100 images selected from COCO validation split. Similar to qualitative results, the higher bit images have higher PSNR and SSIM values, indicating better image quality. Additionally, the expanded images produced by our method have significantly higher PSNR and SSIM values compared to the corresponding compressed ones. It is clear that our method can effectively improve the image quality especially for lower bit depth images.

\begin{figure*}[!tb]
\centering
  \includegraphics[width=17cm]{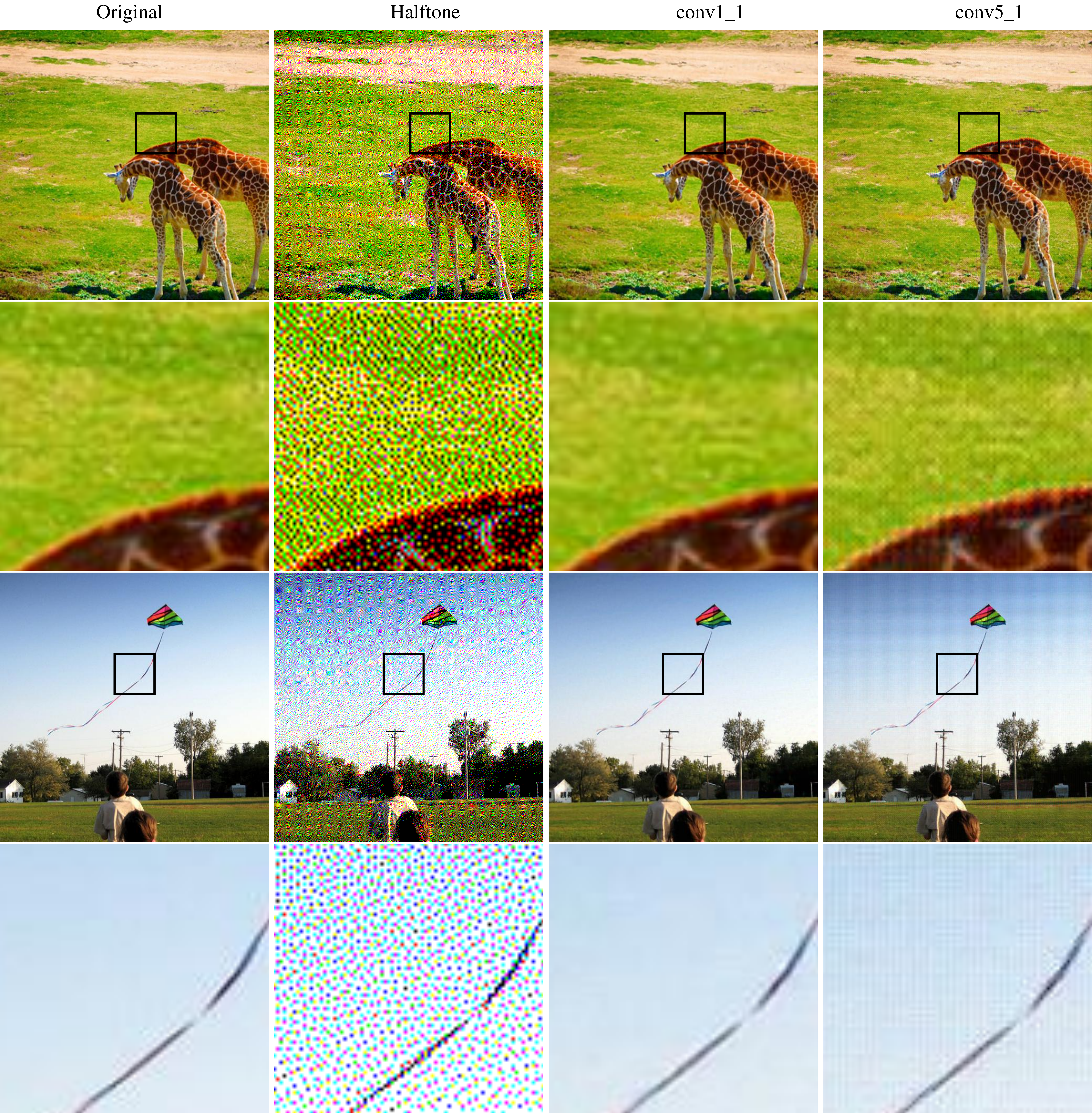}
  \caption{Inverse halftoning results on images from Microsoft COCO validation split. The conv1\_1, conv5\_1 are the results produced by the models trained by the perceptual losses of corresponding convolutional layers. Additionally an enlarged sub-image of each image is given at the bottom for better comparison.}
  \label{fig:halftone}
\end{figure*}

\subsubsection{Perceptual Loss at Different Convolutional Layers}
Due to the multi-layer architecture of deep convolutional neural networks, the perceptual loss can be defined by different convolutional layers. Therefore, we conduct experiments to investigate the performance for different perceptual losses. In all the experiment we use 3 bit depths (8 color levels) as input images to train our deep networks for both color and grayscale images.

\textbf{Qualitative Results.}
As shown in Fig. \ref{fig:depth3_companding}, all the expanded outputs can effectively reduce the blocking and contouring artifacts and reveal continuous-tone results in general. However, the reconstruction based on perceptual loss of higher-level layers could introduce new artifacts such as grid patterns as shown in images for conv3\_1 and conv5\_1 layers. We observed similar phenomenons for input images of different bit-depths. One explanation to this is that, the higher layers will cover a larger area in the input image, and the areas covered by conv3\_1 and conv5\_1 layers are too large to construct a natural looking image. That is, spatial correlations across a large area of the image do not capture natural appearances of an image. Expanding based on perceptual loss of deep features of conv1\_1 layer or lower layers does not have this kind of artifacts. This could be also validated by previous work \cite{mahendran2015understanding} that tries to compute an approximate inverse image from its deep features. It shows that the first few layers in a pretrained CNN are essentially an invertible code of the image and maintain a photographically faithful representations, and the higher level features are corresponding to a more coarse space area of the encoded image.


\textbf{Quantitative Results.}
Table \ref{table:psnr_ssim_depth3_companding} shows the average PSNR and SSIM values for 100 COCO testing images based on perceptual losses constructed with different convolutional layers. On the one hand, both the PSNR and SSIM of our expanded images are much higher than those of the compressed lower bit images, and the compressed images can be significantly improved by our method. On the other hand, the expanded images based on perceptual losses of lower level layers have higher PSNR and SSIM values. This is because new artifacts like grid pattern will be introduced (Fig. \ref{fig:depth3_companding}) although the blocking artifacts can be reduced.



\begin{table}[]
\scriptsize
\centering
\caption{The average companding results of PSNR(dB) and SSIM for 100 color and grayscale testing images. The compressed input images are 3 bits, and the expanded results based on perceptual loss constructed with different convolutional layers are shown.}
\label{table:psnr_ssim_depth3_companding}
\begin{tabular}{c|c|c|c|c|c}
\hline
\multicolumn{2}{c|}{\multirow{2}{*}{\begin{tabular}[c]{@{}c@{}}Perceptual\\ Loss Layer\end{tabular}}} & \multicolumn{2}{c|}{PSNR} & \multicolumn{2}{c}{SSIM} \\ \cline{3-6}
\multicolumn{2}{c|}{}                                                                                & Compressed     & Expanded     & Compressed     & Expanded     \\ \hline
\multirow{2}{*}{Conv1}                                   & Color                                      & 23.13          & 32.64    & 0.85           & 0.93     \\ \cline{2-6}
                                                         & Grayscale                                  & 23.16          & 32.57    & 0.78           & 0.91     \\ \hline
\multirow{2}{*}{Conv2}                                   & Color                                      & 23.13          & 30.00    & 0.85           & 0.88     \\ \cline{2-6}
                                                         & Grayscale                                  & 23.16          & 30.74    & 0.78           & 0.85     \\ \hline
\multirow{2}{*}{Conv3}                                   & Color                                      & 23.13          & 28.17    & 0.85           & 0.87     \\ \cline{2-6}
                                                         & Grayscale                                  & 23.16          & 29.60    & 0.78           & 0.81     \\ \hline
\multirow{2}{*}{Conv4}                                   & Color                                      & 23.13          & 25.74    & 0.85           & 0.84     \\ \cline{2-6}
                                                         & Grayscale                                  & 23.16          & 29.54    & 0.78           & 0.82     \\ \hline
\multirow{2}{*}{Conv5}                                   & Color                                      & 23.13          & 25.43    & 0.85           & 0.87     \\ \cline{2-6}
                                                         & Grayscale                                  & 23.16          & 27.50    & 0.78           & 0.77     \\ \hline
\end{tabular}
\end{table}

\begin{figure*}[!htb]
\centering
  \includegraphics[width=17cm]{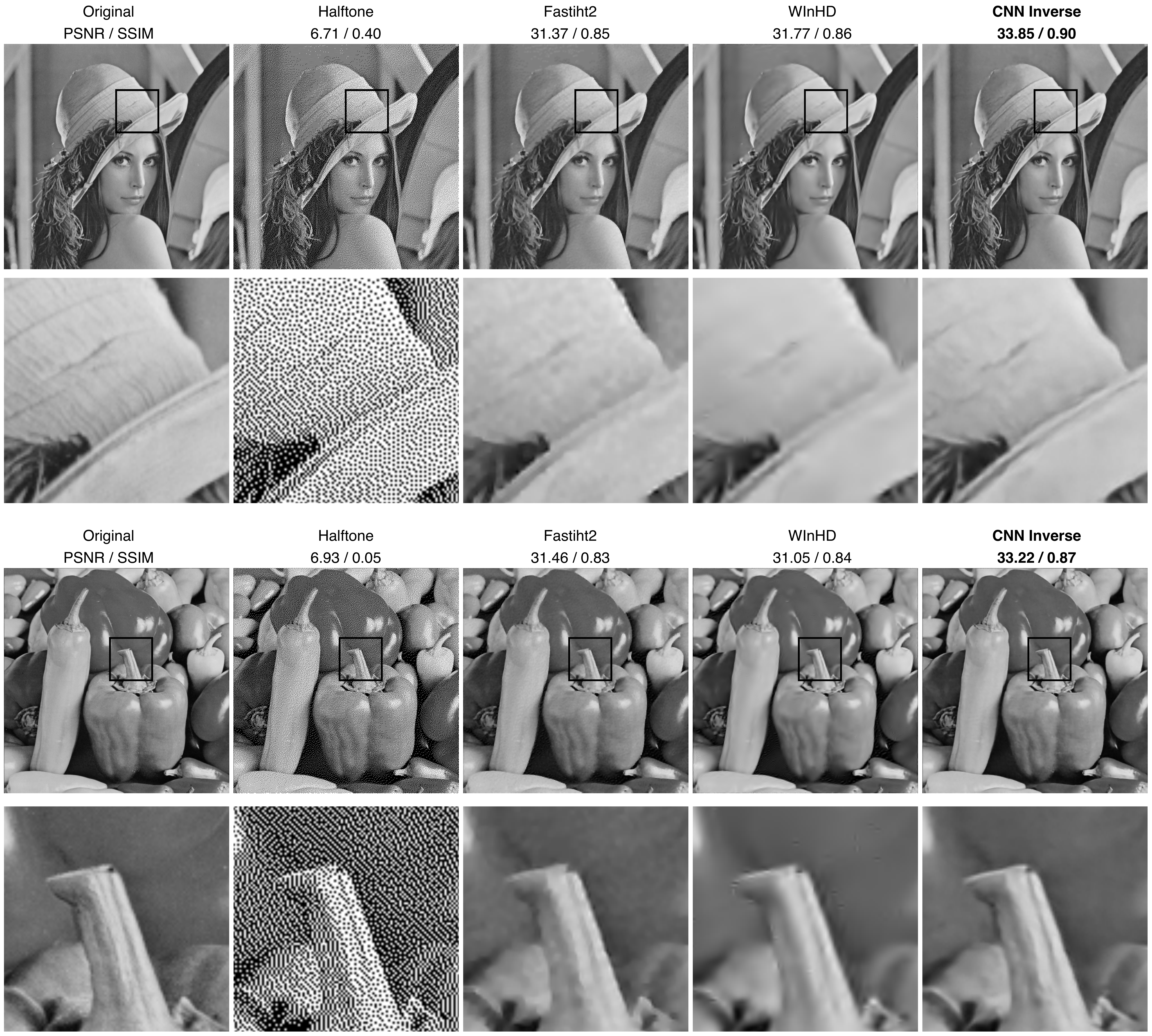}
  \caption{A comparison of inverse halftoning results on grayscale \textit{Lena} and \textit{Peppers} images by different methods. We compare our CNN Inverse method with those of Fastiht2 \cite{kite2000fast} and Wavelet-based WInHD \cite{neelamani2002winhd}. We report PSNR / SSIM for each example.}
  \label{fig:other_hafltone}
\end{figure*}

\begin{figure*}[!htb]
\centering
  \includegraphics[width=17cm]{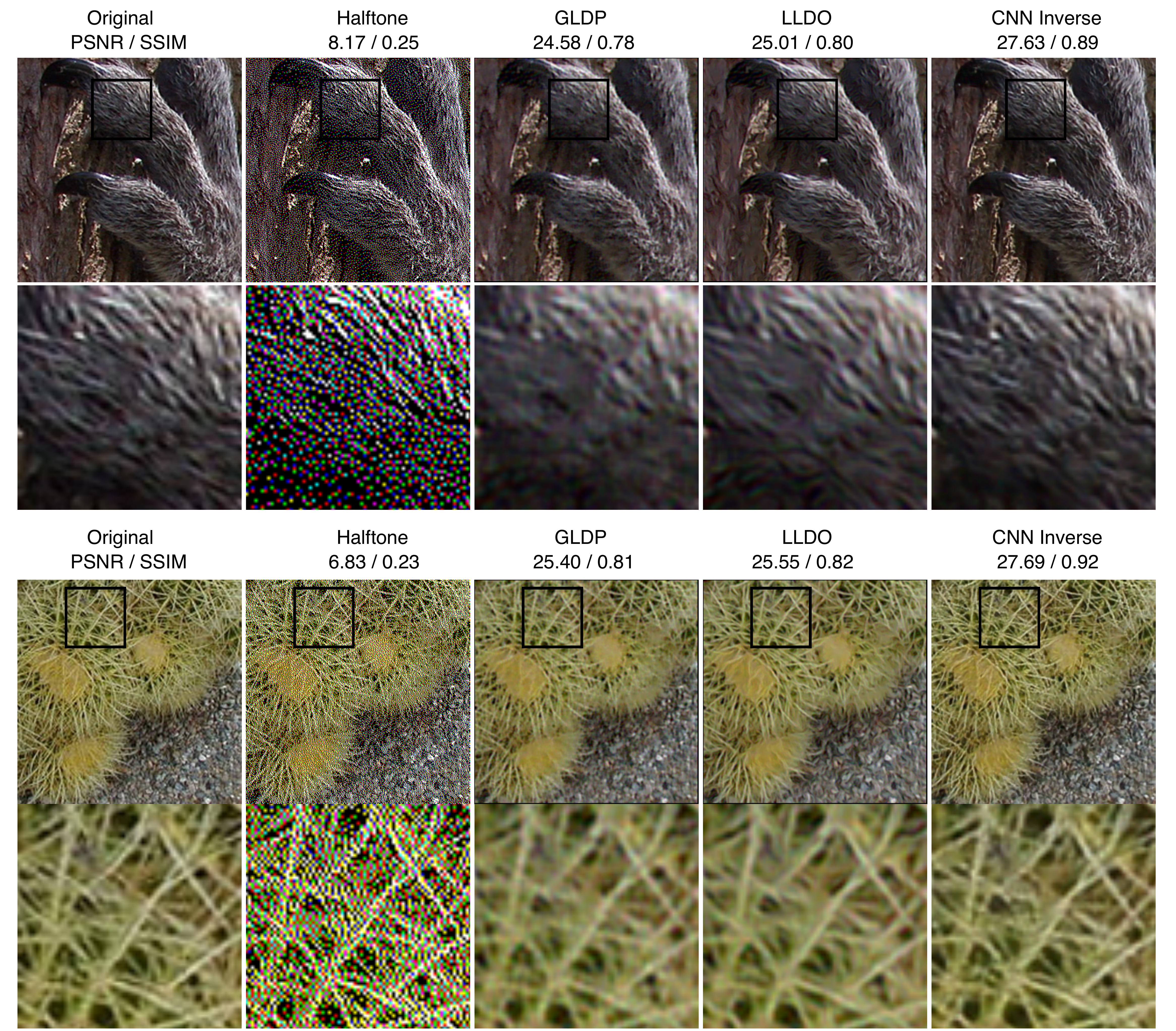}
  \caption{A comparison of inverse halftoning results on color \textit{Koala} and \textit{Cactus} images by different methods. We compare our CNN Inverse method with those of GLDP \cite{son2012inverse} and LLDO \cite{son2014local}. We report PSNR / SSIM for each example.}
  \label{fig:lena_color_halftone}
\end{figure*}

\subsection{Inverse halftoning}
Another similar image processing problem we are interested in is inverse halftoning. This task is to generate a continuous-tone image from halftoned binary images. This problem is also inherently ill-posed since there could exist multiple continuous-tone images corresponding to the halftoned ones. In our experiments we try to use deep feature based perceptual loss to allow the inversed halftones perceptually similar to the given targets. We experiment with both color and grayscale images by using the same approach and employ error diffusion based Floyd–Steinberg dithering for halftoning.

\textbf{Qualitative Results.}
We test our models on random samples of images from Microsoft COCO validation split. The inverse halftoning results are shown in Fig. \ref{fig:halftone}. We can see that the inversed outputs produced by our method are visually similar to the original images. All the outputs can show much smoother textures and produce sharper edges. For instance, sharp kite line and smooth sky can be reconstructed in the kite image. When comparing with the inversed outputs produced by using perceptual loss of different level layers, the outputs from lower-level layer is visually better than those from higher-level layer. Like image companding, grid pattern artifacts can be introduced when using higher-level layer to construct perceptual loss.

\begin{table}[]
\scriptsize
\centering
\caption{The average inversed halftoning results of PSNR(dB) and SSIM for 100 color and grayscale images selected from Microsoft COCO validation split.}
\label{table:halftone}
\begin{tabular}{c|c|c|c|c|c}
\hline
\multicolumn{2}{c|}{\multirow{2}{*}{\begin{tabular}[c]{@{}c@{}}Perceptual\\ Loss Layer\end{tabular}}} & \multicolumn{2}{c|}{PSNR} & \multicolumn{2}{c}{SSIM} \\ \cline{3-6}
\multicolumn{2}{c|}{}                                                                               & Halftone      & CNN Inverse      & Halftone      & CNN Inverse      \\ \hline
\multirow{2}{*}{Conv1}                                  & Color                                      & 8.08          & 31.43     & 0.20          & 0.91      \\ \cline{2-6}
                                                        & Grayscale                                  & 7.92          & 31.36     & 0.14          & 0.90      \\ \hline
\multirow{2}{*}{Conv2}                                  & Color                                      & 8.08          & 20.98     & 0.20          & 0.59      \\ \cline{2-6}
                                                        & Grayscale                                  & 7.92          & 23.98     & 0.14          & 0.67      \\ \hline
\multirow{2}{*}{Conv3}                                  & Color                                      & 8.08          & 24.05     & 0.20          & 0.73      \\ \cline{2-6}
                                                        & Grayscale                                  & 7.92          & 27.44     & 0.14          & 0.74      \\ \hline
\multirow{2}{*}{Conv4}                                  & Color                                      & 8.08          & 26.48     & 0.20          & 0.85      \\ \cline{2-6}
                                                        & Grayscale                                  & 7.92          & 27.82     & 0.14          & 0.76      \\ \hline
\multirow{2}{*}{Conv5}                                  & Color                                      & 8.08          & 25.47     & 0.20          & 0.84      \\ \cline{2-6}
                                                        & Grayscale                                  & 7.92          & 26.48     & 0.14          & 0.69      \\ \hline
\end{tabular}
\end{table}

In addition, we also compare our method on two widely used grayscale images \textit{Lenna} and \textit{Peppers} with other algorithms. Fig. \ref{fig:other_hafltone} shows comparative grayscale results against previous Fastiht2 \cite{kite2000fast} and Wavelet-based WInHD \cite{neelamani2002winhd} algorithms. We also report the PSNR / SSIM measurement for each image. It is clear that our learning-based method can achieve state-of-the-art results and produce sharp edges and fine details, such as the hat in the Lenna image. Our deep models can effectively and correctly learn the relevant spatial correlation and semantic between different pixels and infer the ``best" values for a single pixel based on its neighbors. Moreover, our method can be naturally adapted to color images and produce high-quality continuous-tone color images from corresponding halftones. Fig. \ref{fig:lena_color_halftone} shows the resulting images for the \textit{Koala} and \textit{Cactus} image, which include fine textures and structures. We compare our results (CNN Inverse) with those of two recent methods GLDP \cite{son2012inverse} and LLDO \cite{son2014local}. We can see that our method can provide better resulting images with well expressed fur and bark in the \textit{Koala} image, and distinct boundaries of the fine sand and sharpened edges of splines in the \textit{Cactus} image.

\begin{table*}[!htb]
\centering
\caption{PSNR (dB) and SSIM comparison of different inverse halftoning methods for color images: MAP \cite{stevenson1997inverse}, ALF \cite{kite2000fast}, LPA-ICI \cite{foi2004inverse}, GLDP \cite{son2012inverse}, LLDO \cite{son2014local} and our CNN Inverse.}
\label{table:color_other_inverse}
\begin{tabular}{c|c|c|c|c|c|c|c|c|c|c|c|c}
\hline
\multirow{2}{*}{Image} & \multicolumn{2}{c|}{ALF} & \multicolumn{2}{c|}{MAP} & \multicolumn{2}{c|}{LPA-ICI} & \multicolumn{2}{c|}{GLDP} & \multicolumn{2}{c|}{LLDO} & \multicolumn{2}{c}{CNN Inverse} \\ \cline{2-13} 
                       & PSNR        & SSIM       & PSNR        & SSIM       & PSNR          & SSIM         & PSNR        & SSIM        & PSNR         & SSIM       & PSNR            & SSIM           \\ \hline
Koala                  & 22.36       & 0.66       & 23.33       & 0.74       & 24.17         & 0.76         & 24.58       & 0.78       & 25.01        & 0.80       & 27.63           & 0.89           \\ \hline
Cactus                 & 22.99       & 0.64       & 23.95       & 0.77       & 25.04         & 0.79         & 25.40       & 0.81        & 25.55        & 0.82       & 27.69           & 0.92           \\ \hline
Bear                   & 21.82       & 0.62       & 22.63       & 0.72       & 23.14         & 0.72         & 23.66       & 0.77        & 24.17        & 0.78       & 26.35           & 0.89           \\ \hline
Barbara                & 25.41       & 0.71       & 26.24       & 0.78       & 27.88         & 0.83         & 27.12       & 0.80        & 28.48        & 0.85       & 31.79           & 0.92           \\ \hline
Shop                   & 22.14       & 0.64       & 22.46       & 0.69       & 24.12         & 0.77         & 23.86       & 0.75        & 24.61        & 0.80       & 27.27           & 0.89           \\ \hline
Peppers                & 30.92       & 0.87       & 28.25       & 0.77       & 30.70         & 0.87         & 30.92       & 0.87        & 31.07        & 0.87       & 31.44           & 0.89           \\ \hline
\end{tabular}
\end{table*}

\textbf{Quantitative Results.}
We use PSNR and SSIM as quality metrics to quantitatively evaluate our inverse halftoning results. Table \ref{table:halftone} shows the average PSNR and SSIM values for 100 COCO testing images constructed from different convolutional layers.
It is clear that based on these image evaluation metrics, our method can improve the images by a large margin for both color and grayscale images. In our experiment, the best results are produced by the model trained with conv1\_1 layer.
When using perceptual loss based on higher layers gives rise to a slight grid pattern artifacts visible under magnification, which harms the PSNR and SSIM.

Moreover, we conduct experiments to compare with several previous methods. We use 6 images \textit{Koala}, \textit{Cactus}, \textit{Bear}, \textit{Barbara}, \textit{Shop} and \textit{Peppers}, the same as \cite{son2014local} for testing. Table \ref{table:color_other_inverse} shows the PSNR and SSIM results for conventional methods based on MAP estimation \cite{stevenson1997inverse}, ALF \cite{kite2000fast}, LPA-ICI \cite{foi2004inverse} and recent GLDP \cite{son2012inverse} and LLDO \cite{son2014local}. We can see that our algorithm (CNN Inverse) can achieve new state-of-the-art results and significantly outperform previous methods for inverse halftoning.

\section{Discussion}
Image companding and inverse halftoning are two similar image processing problems in the sense that they attempt to use a lower bit depth image to represent a higher bit depth version of the same image. The naive bit depth compression in image companding is directly applying image quantization technique. It can retain the overall structure and color contrast, however blocking and contouring artifacts will be introduced that make the compressed images look unnatural with visually annoying artifacts. Halftone images try to simulate continuous-tone imagery through the use of dots with only two color levels per channel. The reproduction of halftones for humans relies on an optical illusion that tiny halftone dots could be blended into smooth tones by human eyes. In order to expand the compressed images and inverse the halftones, traditional methods usually need to design expanding and inverse operators manually. For example, the halftone technique such as the specific dithering algorithms should be given in advance in order to design an inverse operator. In this paper, we show that a learning based method can formulate the two problems in the same framework and a perceptual loss based on pretrained deep networks can be used to guide the training. This paper demonstrates that deep convolutional neural networks can not only be applied to high-level vision problems like image classification, but also to traditional low-level vision problems. Although we can use popular metrics like PSNR and SSIM to quantitatively measure the image quality, it is worth pointing out that the assessment of image quality is still a challenging problem. PSNR and SSIM could correlate poorly with human assessment of visual quality and further works are needed for perceptually better image measurement.

\section{Conclusion}
In this paper, we propose to train deep convolutional neural networks with a perceptual loss for two low-level image processing problems: image companding and inverse halftoning. Our method is very effective in dealing with compressed blocking and contouring artifacts for companding and reproduces state-of-the-art continuous-tone outputs from binary halftone images. In addition, we systematically investigated how the perceptual loss constructed with different convolutional layers of the pretrained deep network affects the generated image quality.

\ifCLASSOPTIONcaptionsoff
  \newpage
\fi



%

\bibliographystyle{IEEEtran}
\bibliography{IEEEabrv,IEEEexample}

\begin{thebibliography}{10}
\providecommand{\url}[1]{#1}
\csname url@samestyle\endcsname
\providecommand{\newblock}{\relax}
\providecommand{\bibinfo}[2]{#2}
\providecommand{\BIBentrySTDinterwordspacing}{\spaceskip=0pt\relax}
\providecommand{\BIBentryALTinterwordstretchfactor}{4}
\providecommand{\BIBentryALTinterwordspacing}{\spaceskip=\fontdimen2\font plus
\BIBentryALTinterwordstretchfactor\fontdimen3\font minus
  \fontdimen4\font\relax}
\providecommand{\BIBforeignlanguage}[2]{{%
\expandafter\ifx\csname l@#1\endcsname\relax
\typeout{** WARNING: IEEEtran.bst: No hyphenation pattern has been}%
\typeout{** loaded for the language `#1'. Using the pattern for}%
\typeout{** the default language instead.}%
\else
\language=\csname l@#1\endcsname
\fi
#2}}
\providecommand{\BIBdecl}{\relax}
\BIBdecl

\bibitem{krizhevsky2012imagenet}
A.~Krizhevsky, I.~Sutskever, and G.~E. Hinton, ``Imagenet classification with
  deep convolutional neural networks,'' in \emph{Advances in neural information
  processing systems}, 2012, pp. 1097--1105.

\bibitem{simonyan2014very}
K.~Simonyan and A.~Zisserman, ``Very deep convolutional networks for
  large-scale image recognition,'' \emph{arXiv preprint arXiv:1409.1556}, 2014.

\bibitem{li2005compressing}
Y.~Li, L.~Sharan, and E.~H. Adelson, ``Compressing and companding high dynamic
  range images with subband architectures,'' in \emph{ACM transactions on
  graphics (TOG)}, vol.~24, no.~3.\hskip 1em plus 0.5em minus 0.4em\relax ACM,
  2005, pp. 836--844.

\bibitem{kite2000fast}
T.~D. Kite, N.~Damera-Venkata, B.~L. Evans, and A.~C. Bovik, ``A fast,
  high-quality inverse halftoning algorithm for error diffused halftones,''
  \emph{IEEE Transactions on Image Processing}, vol.~9, no.~9, pp. 1583--1592,
  2000.

\bibitem{shen2001robust}
M.-Y. Shen and C.-C.~J. Kuo, ``A robust nonlinear filtering approach to inverse
  halftoning,'' \emph{Journal of Visual Communication and Image
  Representation}, vol.~12, no.~1, pp. 84--95, 2001.

\bibitem{easley2009inverse}
G.~R. Easley, V.~M. Patel, and D.~M. Healy~Jr, ``Inverse halftoning using a
  shearlet representation,'' in \emph{SPIE Optical Engineering+
  Applications}.\hskip 1em plus 0.5em minus 0.4em\relax International Society
  for Optics and Photonics, 2009, pp. 74\,460C--74\,460C.

\bibitem{mahendran2015understanding}
A.~Mahendran and A.~Vedaldi, ``Understanding deep image representations by
  inverting them,'' in \emph{Proceedings of the IEEE conference on computer
  vision and pattern recognition}, 2015, pp. 5188--5196.

\bibitem{yang2004integer}
B.~Yang, M.~Schmucker, W.~Funk, C.~Busch, and S.~Sun, ``Integer dct-based
  reversible watermarking for images using companding technique,'' in
  \emph{Electronic Imaging 2004}.\hskip 1em plus 0.5em minus 0.4em\relax
  International Society for Optics and Photonics, 2004, pp. 405--415.

\bibitem{bhooshan20102d}
S.~Bhooshan, V.~Kumar, and H.~Solan, ``2d t-law: a novel approach for image
  companding,'' in \emph{Proceedings of the 4th WSEAS international conference
  on Circuits, systems, signal and telecommunications}.\hskip 1em plus 0.5em
  minus 0.4em\relax World Scientific and Engineering Academy and Society
  (WSEAS), 2010, pp. 19--22.

\bibitem{ting1994error}
M.~Y. Ting and E.~A. Riskin, ``Error-diffused image compression using a
  binary-to-gray-scale decoder and predictive pruned tree-structured vector
  quantization,'' \emph{IEEE Transactions on Image Processing}, vol.~3, no.~6,
  pp. 854--858, 1994.

\bibitem{unal2001restoration}
G.~B. Unal and A.~E. {\c{C}}etin, ``Restoration of error-diffused images using
  projection onto convex sets,'' \emph{IEEE Transactions on Image Processing},
  vol.~10, no.~12, pp. 1836--1841, 2001.

\bibitem{stevenson1997inverse}
R.~L. Stevenson, ``Inverse halftoning via map estimation,'' \emph{IEEE
  Transactions on Image Processing}, vol.~6, no.~4, pp. 574--583, 1997.

\bibitem{neelamani2002winhd}
R.~Neelamani, R.~D. Nowak, and R.~G. Baraniuk, ``Winhd: Wavelet-based inverse
  halftoning via deconvolution,'' \emph{IEEE Transactions on Image Processing},
  2002.

\bibitem{liu2011inverse}
Y.-F. Liu, J.-M. Guo, and J.-D. Lee, ``Inverse halftoning based on the bayesian
  theorem,'' \emph{IEEE Transactions on Image Processing}, vol.~20, no.~4, pp.
  1077--1084, 2011.

\bibitem{minami2012inverse}
Y.~Minami, S.-i. Azuma, and T.~Sugie, ``Inverse halftoning using
  super-resolution image processing,'' \emph{IEEJ Transactions on Electrical
  and Electronic Engineering}, vol.~7, no.~2, pp. 208--213, 2012.

\bibitem{mese2001look}
M.~Mese and P.~P. Vaidyanathan, ``Look-up table (lut) method for inverse
  halftoning,'' \emph{IEEE Transactions on Image Processing}, vol.~10, no.~10,
  pp. 1566--1578, 2001.

\bibitem{son2012inverse}
C.-H. Son, ``Inverse halftoning based on sparse representation,'' \emph{Optics
  letters}, vol.~37, no.~12, pp. 2352--2354, 2012.

\bibitem{son2014local}
C.-H. Son and H.~Choo, ``Local learned dictionaries optimized to edge
  orientation for inverse halftoning,'' \emph{IEEE Transactions on Image
  Processing}, vol.~23, no.~6, pp. 2542--2556, 2014.

\bibitem{freitas2016enhancing}
P.~G. Freitas, M.~C. Farias, and A.~P. Ara{\'u}jo, ``Enhancing inverse
  halftoning via coupled dictionary training,'' \emph{Signal Processing: Image
  Communication}, vol.~49, pp. 1--8, 2016.

\bibitem{long2015fully}
J.~Long, E.~Shelhamer, and T.~Darrell, ``Fully convolutional networks for
  semantic segmentation,'' in \emph{Proceedings of the IEEE Conference on
  Computer Vision and Pattern Recognition}, 2015, pp. 3431--3440.

\bibitem{eigen2015predicting}
D.~Eigen and R.~Fergus, ``Predicting depth, surface normals and semantic labels
  with a common multi-scale convolutional architecture,'' in \emph{Proceedings
  of the IEEE International Conference on Computer Vision}, 2015, pp.
  2650--2658.

\bibitem{noh2015learning}
H.~Noh, S.~Hong, and B.~Han, ``Learning deconvolution network for semantic
  segmentation,'' in \emph{Proceedings of the IEEE International Conference on
  Computer Vision}, 2015, pp. 1520--1528.

\bibitem{iizuka2016let}
S.~Iizuka, E.~Simo-Serra, and H.~Ishikawa, ``Let there be color!: joint
  end-to-end learning of global and local image priors for automatic image
  colorization with simultaneous classification,'' \emph{ACM Transactions on
  Graphics (TOG)}, vol.~35, no.~4, p. 110, 2016.

\bibitem{larsson2016learning}
G.~Larsson, M.~Maire, and G.~Shakhnarovich, ``Learning representations for
  automatic colorization,'' in \emph{European Conference on Computer
  Vision}.\hskip 1em plus 0.5em minus 0.4em\relax Springer, 2016, pp. 577--593.

\bibitem{eigen2014depth}
D.~Eigen, C.~Puhrsch, and R.~Fergus, ``Depth map prediction from a single image
  using a multi-scale deep network,'' in \emph{Advances in neural information
  processing systems}, 2014, pp. 2366--2374.

\bibitem{liu2015deep}
F.~Liu, C.~Shen, and G.~Lin, ``Deep convolutional neural fields for depth
  estimation from a single image,'' in \emph{Proceedings of the IEEE Conference
  on Computer Vision and Pattern Recognition}, 2015, pp. 5162--5170.

\bibitem{xie2015holistically}
S.~Xie and Z.~Tu, ``Holistically-nested edge detection,'' in \emph{Proceedings
  of the IEEE International Conference on Computer Vision}, 2015, pp.
  1395--1403.

\bibitem{wang2004image}
Z.~Wang, A.~C. Bovik, H.~R. Sheikh, and E.~P. Simoncelli, ``Image quality
  assessment: from error visibility to structural similarity,'' \emph{IEEE
  transactions on image processing}, vol.~13, no.~4, pp. 600--612, 2004.

\bibitem{chen2014semantic}
L.-C. Chen, G.~Papandreou, I.~Kokkinos, K.~Murphy, and A.~L. Yuille, ``Semantic
  image segmentation with deep convolutional nets and fully connected crfs,''
  \emph{arXiv preprint arXiv:1412.7062}, 2014.

\bibitem{goodfellow2014generative}
I.~Goodfellow, J.~Pouget-Abadie, M.~Mirza, B.~Xu, D.~Warde-Farley, S.~Ozair,
  A.~Courville, and Y.~Bengio, ``Generative adversarial nets,'' in
  \emph{Advances in neural information processing systems}, 2014, pp.
  2672--2680.

\bibitem{mirza2014conditional}
M.~Mirza and S.~Osindero, ``Conditional generative adversarial nets,''
  \emph{arXiv preprint arXiv:1411.1784}, 2014.

\bibitem{reed2016generative}
S.~Reed, Z.~Akata, X.~Yan, L.~Logeswaran, B.~Schiele, and H.~Lee, ``Generative
  adversarial text to image synthesis,'' in \emph{Proceedings of The 33rd
  International Conference on Machine Learning}, vol.~3, 2016.

\bibitem{li2016precomputed}
C.~Li and M.~Wand, ``Precomputed real-time texture synthesis with markovian
  generative adversarial networks,'' in \emph{European Conference on Computer
  Vision}.\hskip 1em plus 0.5em minus 0.4em\relax Springer, 2016, pp. 702--716.

\bibitem{pathak2016context}
D.~Pathak, P.~Krahenbuhl, J.~Donahue, T.~Darrell, and A.~A. Efros, ``Context
  encoders: Feature learning by inpainting,'' in \emph{Proceedings of the IEEE
  Conference on Computer Vision and Pattern Recognition}, 2016, pp. 2536--2544.

\bibitem{mathieu2015deep}
M.~Mathieu, C.~Couprie, and Y.~LeCun, ``Deep multi-scale video prediction
  beyond mean square error,'' \emph{arXiv preprint arXiv:1511.05440}, 2015.

\bibitem{isola2016image}
P.~Isola, J.-Y. Zhu, T.~Zhou, and A.~A. Efros, ``Image-to-image translation
  with conditional adversarial networks,'' \emph{arXiv preprint
  arXiv:1611.07004}, 2016.

\bibitem{yosinski2015understanding}
J.~Yosinski, J.~Clune, A.~Nguyen, T.~Fuchs, and H.~Lipson, ``Understanding
  neural networks through deep visualization,'' \emph{arXiv preprint
  arXiv:1506.06579}, 2015.

\bibitem{simonyan2013deep}
K.~Simonyan, A.~Vedaldi, and A.~Zisserman, ``Deep inside convolutional
  networks: Visualising image classification models and saliency maps,''
  \emph{arXiv preprint arXiv:1312.6034}, 2013.

\bibitem{dosovitskiy2016inverting}
A.~Dosovitskiy and T.~Brox, ``Inverting visual representations with
  convolutional networks,'' in \emph{Proceedings of the IEEE Conference on
  Computer Vision and Pattern Recognition}, 2016, pp. 4829--4837.

\bibitem{gatys2015neural}
L.~A. Gatys, A.~S. Ecker, and M.~Bethge, ``A neural algorithm of artistic
  style,'' \emph{arXiv preprint arXiv:1508.06576}, 2015.

\bibitem{gatys2015texture}
L.~Gatys, A.~S. Ecker, and M.~Bethge, ``Texture synthesis using convolutional
  neural networks,'' in \emph{Advances in Neural Information Processing
  Systems}, 2015, pp. 262--270.

\bibitem{johnson2016perceptual}
J.~Johnson, A.~Alahi, and L.~Fei-Fei, ``Perceptual losses for real-time style
  transfer and super-resolution,'' in \emph{European Conference on Computer
  Vision}.\hskip 1em plus 0.5em minus 0.4em\relax Springer, 2016, pp. 694--711.

\bibitem{hou2017deep}
X.~Hou, L.~Shen, K.~Sun, and G.~Qiu, ``Deep feature consistent variational
  autoencoder,'' in \emph{Applications of Computer Vision (WACV), 2017 IEEE
  Winter Conference on}.\hskip 1em plus 0.5em minus 0.4em\relax IEEE, 2017, pp.
  1133--1141.

\bibitem{he2016deep}
K.~He, X.~Zhang, S.~Ren, and J.~Sun, ``Deep residual learning for image
  recognition,'' in \emph{Proceedings of the IEEE Conference on Computer Vision
  and Pattern Recognition}, 2016, pp. 770--778.

\bibitem{ronneberger2015u}
O.~Ronneberger, P.~Fischer, and T.~Brox, ``U-net: Convolutional networks for
  biomedical image segmentation,'' in \emph{International Conference on Medical
  Image Computing and Computer-Assisted Intervention}.\hskip 1em plus 0.5em
  minus 0.4em\relax Springer, 2015, pp. 234--241.

\bibitem{radford2015unsupervised}
A.~Radford, L.~Metz, and S.~Chintala, ``Unsupervised representation learning
  with deep convolutional generative adversarial networks,'' \emph{arXiv
  preprint arXiv:1511.06434}, 2015.

\bibitem{ioffe2015batch}
S.~Ioffe and C.~Szegedy, ``Batch normalization: Accelerating deep network
  training by reducing internal covariate shift,'' \emph{arXiv preprint
  arXiv:1502.03167}, 2015.

\bibitem{collobert2011torch7}
R.~Collobert, K.~Kavukcuoglu, and C.~Farabet, ``Torch7: A matlab-like
  environment for machine learning,'' in \emph{BigLearn, NIPS Workshop}, no.
  EPFL-CONF-192376, 2011.

\bibitem{kingma2014adam}
D.~Kingma and J.~Ba, ``Adam: A method for stochastic optimization,''
  \emph{arXiv preprint arXiv:1412.6980}, 2014.

\bibitem{lin2014microsoft}
T.-Y. Lin, M.~Maire, S.~Belongie, J.~Hays, P.~Perona, D.~Ramanan,
  P.~Doll{\'a}r, and C.~L. Zitnick, ``Microsoft coco: Common objects in
  context,'' in \emph{European Conference on Computer Vision}.\hskip 1em plus
  0.5em minus 0.4em\relax Springer, 2014, pp. 740--755.

\bibitem{foi2004inverse}
A.~Foi, V.~Katkovnik, K.~Egiazarian, and J.~Astola, ``Inverse halftoning based
  on the anisotropic lpa-ici deconvolution,'' in \emph{Proceedings of Int.
  TICSP Workshop Spectral Meth. Multirate Signal Process}, 2004.

\end{thebibliography}

%








\end{document}